\documentclass[
]{ceurart}

\usepackage{listings}
\usepackage{float} 

\usepackage{booktabs}
\usepackage{multirow}
\usepackage{makecell}

\usepackage{threeparttable}

\begin{document}

\copyrightyear{2026}
\copyrightclause{Copyright for this paper by its authors.
  Use permitted under Creative Commons License Attribution 4.0
  International (CC BY 4.0).}

\conference{IRCDL'25: 22nd Conference on Information and Research Science Connecting to Digital and Library Science,
February 19-20 2026, Modena, Italy}

\title{Diagnosing Structural Failures in LLM-Based Evidence Extraction for Meta-Analysis}


\author[1]{Zhiyin Tan}[%
orcid=0009-0002-4166-5810,
email=zhiyin.tan@l3s.de,
]
\cormark[1]
\address[1]{L3S Research Center, Leibniz University Hannover, Hannover, Germany}
\author[2]{Jennifer D'Souza}[%
orcid=0000-0002-6616-9509,
email=jennifer.dsouza@tib.eu,
]
\address[2]{TIB Leibniz Information Centre for Science and Technology, Hannover, Germany}
\cortext[1]{Corresponding author.}

\begin{abstract}
Systematic reviews and meta-analyses rely on converting narrative articles into structured, numerically grounded study records. Despite rapid advances in large language models (LLMs), it remains unclear whether they can meet the structural requirements of this process, which hinge on preserving roles, methods, and effect-size attribution across documents rather than on recognizing isolated entities. We propose a structural, diagnostic framework that evaluates LLM-based evidence extraction as a progression of schema-constrained queries with increasing relational and numerical complexity, enabling precise identification of failure points beyond atom-level extraction. Using a manually curated corpus spanning five scientific domains, together with a unified query suite and evaluation protocol, we evaluate two state-of-the-art LLMs under both per-document and long-context, multi-document input regimes. Across domains and models, performance remains moderate for single-property queries but degrades sharply once tasks require stable binding between variables, roles, statistical methods, and effect sizes. Full meta-analytic association tuples are extracted with near-zero reliability, and long-context inputs further exacerbate these failures. Downstream aggregation amplifies even minor upstream errors, rendering corpus-level statistics unreliable. Our analysis shows that these limitations stem not from entity recognition errors, but from systematic structural breakdowns, including role reversals, cross-analysis binding drift, instance compression in dense result sections, and numeric misattribution, indicating that current LLMs lack the structural fidelity, relational binding, and numerical grounding required for automated meta-analysis. The code and data are publicly available at GitHub.~\footnote{\url{https://github.com/zhiyintan/LLM-Meta-Analysis}}
\end{abstract}


\maketitle

\section{Introduction}

Systematic reviews and meta-analyses are foundational to evidence synthesis across the sciences, providing structured methods for integrating empirical results at scale~\cite{cochrane2019handbook,cooper2015research}. Despite advances in digital libraries and scholarly infrastructures, systematic review workflows remain dominated by manual data extraction, where study-level elements, such as populations, variables, measurement units, association methods, and effect sizes, must be identified, normalized, and encoded for analysis~\cite{tsafnat2014systematic}. This step is highly labor-intensive and remains a major bottleneck in creating reproducible, machine-actionable meta-analytic datasets.

Parallel work in semantic publishing and structured scholarly knowledge has demonstrated that scientific studies can be represented through explicit schemas and ontologies. Examples range from publication-centric models such as SPAR \cite{peroni2018spar}, to domain-specific controlled vocabularies in biomedicine (e.g., MeSH~\cite{lipscomb2000medical}, GO~\cite{ashburner2000gene}, UMLS~\cite{bodenreider2004unified}), and large-scale scholarly knowledge graphs such as Semantic Scholar’s Literature Graph \cite{ammar2018S2LiteratureGraph} and the Computer Science KG \cite{dessi2025cskg}. The Open Research Knowledge Graph (ORKG) \cite{auer2025open} further shows how paper-level claims and methodological details can be captured via human-authored semantic templates.
Large language models (LLMs) appear promising for automating parts of this process. They show strong performance on isolated extraction tasks and domain-adapted scientific text processing \cite{shamsabadi2024large,dagdelen2024structured,lu2025finetuning,d2025mining}, and recent work has explored their use in systematic review screening and semi-automated data extraction pipelines \cite{wang2024autosurvey,guo2025sgsimeval}. However, a growing body of evidence highlights limitations that are directly relevant to the structural demands of meta-analytic workflows. LLMs struggle with compositional generalization, the systematic combination of familiar elements into novel but valid structures \cite{Dziri2023Neural,ismayilzada2025LLMMorphological}. They also exhibit failures in entity and attribute binding, where recognized elements become misaligned or incorrectly associated \cite{feng2024LLMBindEntities,gurarieh2025LLMRetrieveBoundEntities}. Long-context processing introduces further fragility, with models often ignoring or misusing information located in the middle of extended inputs \cite{liu2024LLMLostInLongContext}. Finally, numerically grounded reasoning remains brittle, particularly when values must be correctly tied to specific variables, units, or statistical methods \cite{akhtar2023LLMNumericalReasoningCapabilities,ahmed2025LLMNumericalReasoningCapabilities,oyesanmi2025LLMReasoning}.
Meta-analysis provides a particularly stringent testbed for these weaknesses. Constructing meta-analytic inputs requires extracting and binding multi-entity, numerically annotated, and method-conditioned tuples, such as population, independent variable, dependent variable, association method, sample size, and effect size, across multiple documents. Success therefore depends not only on recognizing individual entities, but on preserving their structural relationships under aggregation.

In this work, we take a structural and diagnostic perspective on LLM-based scientific evidence extraction. Recent work has shown that LLMs can be used for semi-automated evidence extraction in systematic reviews, particularly in biomedical settings~\cite{li2025automated}. Rather than proposing a new extraction model or end-to-end pipeline, we study how current LLMs behave when evidence extraction is framed as a sequence of increasingly structured, schema-constrained tasks.
This framing allows us to explicitly control the relational and binding demands imposed on the model, ranging from isolated entity extraction to higher-order association structures characteristic of meta-analytic inputs.
We further consider lightweight aggregation operations to probe whether extracted evidence can be coherently integrated at the corpus level.
We instantiate this perspective using a generalized semantic study schema and evaluate it across five scientific domains under controlled multi-document settings.
By combining full-text inputs with human-curated gold annotations, our evaluation enables a fine-grained analysis of how model behavior evolves as structural complexity increases. 

This study \textbf{makes four main contributions}:
\begin{enumerate}
    \item a structural and diagnostic perspective on LLM-based scientific evidence extraction, which frames extraction as a progression of increasingly constrained, schema-guided tasks rather than as an end-to-end pipeline problem;
    \item a tiered query framework that operationalizes this perspective through controlled variations in tuple arity and relational binding complexity, covering both direct extraction and derived statistical reasoning;
    \item an manually curated, multi-domain, schema-aligned evaluation benchmark spanning five scientific domains (civil engineering, medical and health science, agricultural science, earth and environmental sciences, and social science) designed to systematically investigate LLM performance under increasing structural demands;
    \item a comprehensive empirical analysis identifying where along this structured progression modern LLMs begin to fail, revealing specific weaknesses in relational binding, numerical grounding, and document-level attribution.
\end{enumerate}

Together, these contributions clarify the structural requirements of automated meta-analytic evidence extraction and provide a foundation for designing future LLM-based systems that better align with the needs of evidence synthesis and digital library infrastructure.

\section{Related Work}

\paragraph{Semantic Modeling and Structured Representations of Scientific Studies.}

Prior work in semantic publishing treats scientific articles as structured objects rather than purely narrative text, modeling key study components such as populations, variables, interventions, outcomes, and measurements using explicit schemas and ontologies~\cite{Peroni2012FaBiOandCiTO,Constantin2016DoCO,Ciccarese2008SWANbiomedicalDiscourseOntology}. This perspective enables machine-interpretable access to scientific content beyond surface-level text.
This idea has been instantiated at scale through ontology suites and scholarly knowledge graphs that formalize document structure, citation relations, and scientific statements~\cite{Peroni2012FaBiOandCiTO,peroni2018spar,Ciccarese2008SWANbiomedicalDiscourseOntology,Kilicoglu2012SemMedDB,Himmelstein2017SystematicIntegrationBiomedical}. The Open Research Knowledge Graph further demonstrates paper-level contribution modeling via human-authored schemas~\cite{oelen2019ORKGcomparing}, while autonomously constructed graphs such as MatKG show that complex scientific entities and relations can be populated at corpus scale~\cite{venugopal2024matkg}. Recent work also explores LLM-assisted schema discovery and refinement~\cite{sadruddin2025llms4schemadiscovery}.
In this setting, a central question is how such structured study representations can be reliably instantiated from unstructured scientific articles at scale when schemas require preserving multi-entity relations and numerically grounded attributes.

\paragraph{Automating Information Extraction in Systematic Reviews and Meta-Analyses.}
Information extraction (IE), often termed data or evidence abstraction in systematic reviews, encodes study-level information into structured forms for downstream synthesis~\cite{cochrane2019handbook,Schmidt2025DataExtractionforReview}. Owing to its cost and error-proneness, IE has been a long-standing target for automation.
Most prior work addresses localized extraction problems, targeting individual elements or simple relations such as populations, interventions, outcomes, or trial characteristics using rule-based, classical machine learning, and neural methods~\cite{Demner-Fushman2007ClinicalQA-KG-Statistical,Wallace2016PICO,Kiritchenko2010ExaCT,Hu2023PICOextraction}. These approaches reduce manual effort but operate primarily on isolated fields.
More recent systems integrate IE into broader review pipelines, including screening, eligibility assessment, and risk-of-bias evaluation, with LLMs increasingly explored as assistants for such tasks~\cite{Brassey2021AutomatedEvidenceSynthesis,Jardim2022RiskOfBiasAutomation,Schmidt2025DataExtractionforReview,Lieberum2025LLMNotReadyforReview,Khan2025CollaborativeLLMsDataExtractionReview}.
Across this literature, task formulations and evaluations remain aligned with low-arity outputs, rather than with the structured, multi-field records required as direct inputs to meta-analysis~\cite{Buscemi2006SingleDataExtractionReviews,Schmidt2025DataExtractionforReview}. In particular, few studies assess whether automated IE preserves the joint binding between populations, variables, methods, and numerically grounded effect measures across documents.
We target this gap by examining LLM-based IE under progressively increasing structural constraints at the pre--meta-analytic stage.

\paragraph{LLMs for Scientific Information Extraction and Multi-Document Reasoning.}

Recent work applies LLMs to scientific information extraction (SciIE), including schema-driven extraction of entities (e.g., tasks, methods, datasets, variables) and relations from full-text articles~\cite{jain2020SciREX,d2025mining, zhang2024sciER,pham2023LabelVariationScientificInfoExtraction,kabongo2024effective}. These tasks often require assembling evidence distributed across sections, rather than extracting isolated sentence-level spans.
In parallel, research on multi-document and multi-hop question answering examines how models integrate evidence across passages or documents. Many approaches construct structured intermediates, such as passage-level or document-level knowledge graphs, to retrieve, align, and compose supporting evidence~\cite{Wang2024KGMulti-DocumentQA,yang2025LLMMulti-DocQAKGReasoning}. Emerging researches spanning scientific articles, citation networks, and multimodal inputs consistently report substantial gaps between LLM and expert performance in multi-document SciIE~\cite{ li2024M3SciQA,T2025MD,park2025Multi-docMulti-hopQA}. Related DocVQA-style benchmarks highlight similar challenges, showing persistent failures in selecting, aligning, and integrating evidence across pages, modalities, and documents, even with retrieval augmentation~\cite{dong2025benchmarkMulti-modelDocumentQA}. Together, these results indicate that long-context ingestion alone is insufficient for reliable cross-document evidence integration.
Most existing formulations nonetheless frame queries as ad-hoc questions or loosely structured prompts, with evaluation centered on answer correctness or evidence selection. Much less is known about whether LLMs can reliably construct schema-conformant, multi-field scientific records that preserve entity bindings and numerically grounded attributes across documents.
We address this gap by grounding all queries in an explicit study schema and systematically varying the arity and structural constraints of the target output, reframing multi-document reasoning as structured evidence assembly rather than question answering.

\paragraph{Compositionality, Binding, and Numeric Reasoning in LLMs.}

A growing body of work documents systematic limitations of LLMs in compositional generalization, entity--attribute binding, and numerically grounded reasoning. Studies on compositionality show that models often fail to generalize to novel combinations of familiar primitives, instead relying on surface patterns rather than systematic structure~\cite{zhao2024CompositionalDeficiencyLLMMathematicalReasoning,Shojaee2025IllusionOfThinking}.
Related work further shows that LLMs struggle to reliably bind attributes, intermediate conclusions, or numeric values to the correct entities, producing fluent but structurally misaligned outputs even when individual facts are known~\cite{song2025LLMReasoningFailures,Mahowald2024LLMLanguageAndThought}. Such binding errors become more pronounced in long-context or multi-document settings, where models may ignore, conflate, or misattribute evidence across sources~\cite{song2025LLMReasoningFailures}.
Numeric reasoning poses additional challenges. Empirical evaluations consistently report brittle performance on arithmetic operations, comparisons, and value grounding, particularly when numeric quantities must be correctly associated with variables, units, or experimental conditions~\cite{boye2025LLMMathematicalReasoningFailures,yan2025LLMElementaryAddition}. These failures reflect not only difficulties with computation, but with preserving structured relationships between numbers and the entities they describe.
These limitations are directly relevant to meta-analytic evidence extraction. Constructing meta-analytic inputs requires composing multiple entities, binding attributes and numeric measures to the correct study components, and maintaining these relationships across documents. As a result, compositionality, binding fidelity, and numeric grounding function not as secondary concerns, but as necessary conditions for producing valid, schema-conformant study records.

\paragraph{Neural--Symbolic and Schema-Aware Approaches to Evidence Synthesis.}

Motivated by limitations of purely neural approaches, a growing line of work explores neural--symbolic and schema-aware methods for scientific information extraction and evidence synthesis. These approaches combine LLMs with explicit symbolic structure, such as ontologies, knowledge graphs, or programmatic constraints, to guide extraction, enforce schema conformity, or support structured reasoning across documents~\cite{Liu2025Neural-SymbolicReasoningKG,Skrlj2026SymbolicNeural,Pan2023LLMsKGs}.
Prior work investigates a range of such integrations, including ontology- or KG-constrained extraction, program-aided pipelines that produce intermediate structured representations for validation and post-processing, and retrieval-augmented or constrained-decoding methods designed to improve fidelity under complex output structures~\cite{zhang2024LLMforKGConstruction,li2025ReasoningOnKGWithChains}. Collectively, these efforts reflect a broader shift toward leveraging explicit structure to improve robustness, interpretability, and aggregation in multi-document scientific settings.
Against this backdrop, an open question is how reliably current LLMs can instantiate structured study representations required for meta-analysis as structural, relational, and numerical demands increase. By holding the target schema fixed, we examine how extraction behavior degrades as evidence assembly becomes more compositional and cross-document, providing a diagnostic perspective that complements existing schema-aware and neural--symbolic approaches. This perspective motivates the evaluation framework introduced in the following section.

\section{Methodology}
\label{sec:method}

\begin{figure}[!htb]
  \centering
  \includegraphics[width=\linewidth]{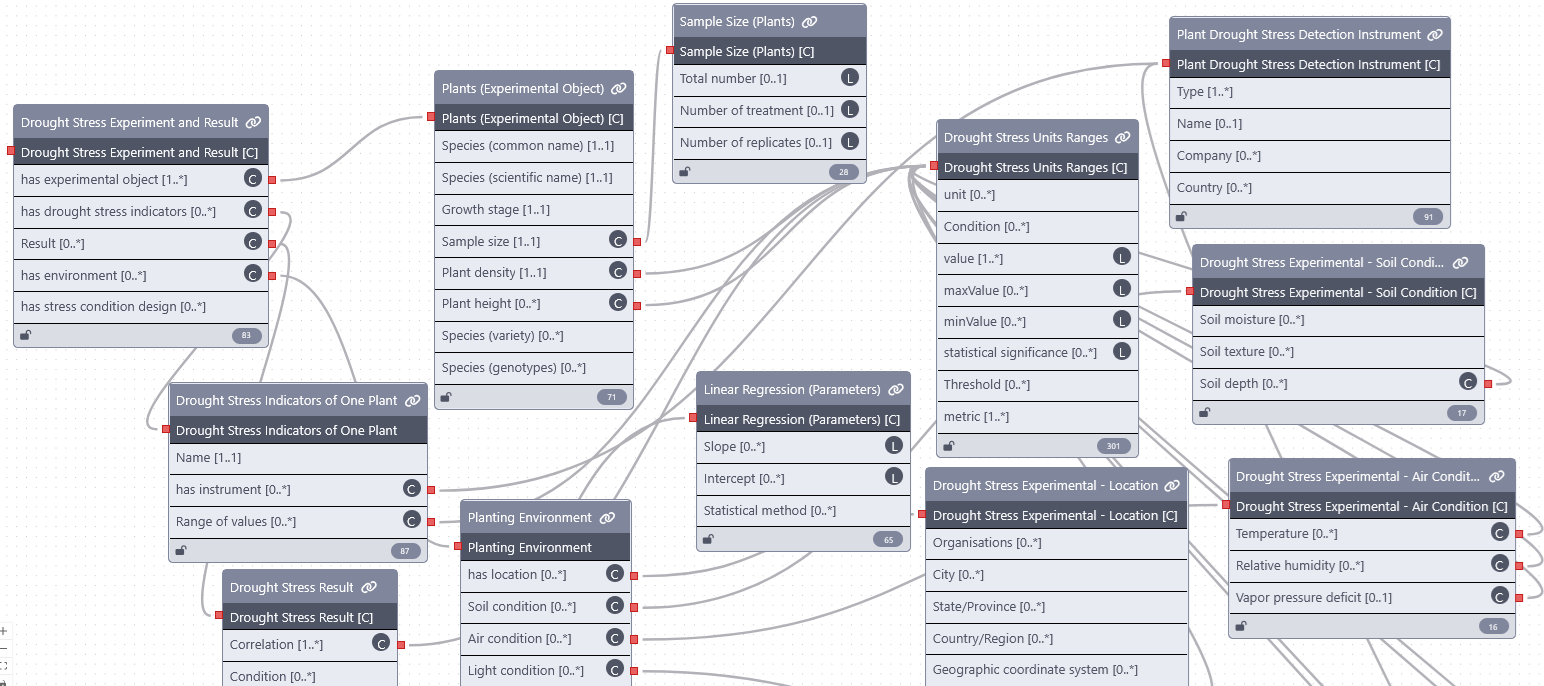}
  \caption{Partial view of the original ORKG template for \textit{Early Drought Stress Indicators in Plants (Physiological \& Multi-Sensor)} \url{https://orkg.org/templates/R1471884}, showing a richly nested, domain-specific schema (plant/specimen descriptors, multi-sensor indicators and instruments, environment/stress conditions, and structured results with value–range–unit measurements) that we generalize, in this study, into a cross-domain study schema for association-based evidence extraction.}
  \label{fig:orkg-semantic-model}
\end{figure}

\subsection{Semantic Study Schema and Application Scope}

Our methodological starting point is a unified semantic representation of empirical studies that supports controlled extraction and aggregation of association-based evidence across domains.
To this end, we adopt a study-level schema that abstracts over discipline-specific reporting practices while retaining the structural elements required for quantitative evidence synthesis.

The schema used in this study builds on an earlier, domain-specific semantic modeling effort within the Open Research Knowledge Graph (ORKG).
That work introduced a highly specialized template (Figure~\ref{fig:orkg-semantic-model}) tailored to a single research question,
and demonstrated how rich, nested semantic structures can support structured cross-paper comparison~\footnote{ORKG template: \url{https://orkg.org/templates/R1471878}, Comparison: \url{https://doi.org/10.48366/R1563383}}. By design, however, both the template and the resulting comparison were domain-specific and not intended to generalize.

In the present work, we substantially simplify and generalize this schema by discarding fine-grained domain semantics and retaining only study elements that recur across association-based empirical research, independent of discipline.
The resulting semantic study schema captures a minimal yet expressive set of entities and relations:
study populations and geographical settings, sample sizes, variables participating in statistical associations, their roles as independent or dependent variables, associated measurement descriptors, statistical methods, and reported effect sizes.
This abstraction deliberately trades domain specificity for cross-domain applicability and structural regularity.
The schema serves as the unifying backbone for all extraction and derivation tasks in our framework, supporting both object-centric queries over study settings and method-centric queries over reported statistical evidence, while consistently preserving document-level attribution.
Within this framework, we focus on a broad but well-structured class of empirical studies commonly used in meta-analysis,
covering effect-size families such as correlation coefficients, standardized regression coefficients, coefficients of determination, and odds ratios~\cite{cochrane2019handbook,borenstein2021IntroMetaAnalysis}.
Concretely, we restrict attention to studies that report explicit statistical associations between variables (e.g., Pearson or Spearman correlations, $\beta$ coefficients, $R$ or $R^2$, OR), which constitute core inputs for effect-size--based synthesis across the behavioral, social, environmental, and agricultural sciences.
This class of studies provides a suitable testbed for structured multi-document evidence extraction: while reporting conventions admit schema-based modeling, correct extraction still requires accurate entity recognition, role assignment, numerical grounding, and consistent document attribution, particularly under many-to-many variable relations.

Table~\ref{tab:schema} summarizes the study-level schema $\mathcal{S}$ used in our evaluation.
Each element corresponds to a schema property that can be instantiated as a document-attributed atom and serve as the basic unit for extraction and derived statistical queries.

\begin{table}[!htb]
\caption{Study-level schema properties used for object- and method-centric evidence extraction. Each schema atom is explicitly indexed by its source document identifier.}
\label{tab:schema}
\centering
\begin{tabular}{p{3.1cm} p{9.2cm} p{2.34cm}}
\toprule
\textbf{Property} & \textbf{Definition} & \textbf{Notation} \\
\midrule
Document ID & Index of the source document in the corpus & $i \in \{1,\ldots,|\mathcal{D}|\}$ \\

\midrule
Study Population & Study population or unit of analysis & $P \in \mathcal{P}$ \\
Geolocation & Country of study conduct (normalize cities/regions to countries) & $G \in \mathcal{G}$ \\
Sample Size & Main reported sample size for the study & $N \in \mathcal{N}$ \\

\midrule
Statistical Method & Statistical or association method (e.g., Pearson, regression, OR) & $A \in \mathcal{A}$ \\
Independent Variable & Predictor variable (role-aware) & $IV \in \mathcal{IV}$ \\
Dependent Variable & Outcome variable (role-aware) & $DV \in \mathcal{DV}$ \\
Variable & Role-agnostic variable (union of $IV$ and $DV$, de-duplicated) & $V \in \mathcal{V}$ \\

\midrule
Scale and Unit & Atomic pair describing measurement scale and unit & $(S,U) \in \mathcal{SU}$ \\
Conditions & Optional contextual qualifier for an association estimate & $C \in \mathcal{C}$ \\
Effect Size & Reported association estimate (e.g., $r$, $\beta$, OR, etc.) & $E \in \mathcal{E}$ \\
\bottomrule
\end{tabular}
\end{table}

\subsection{Problem Formulation}

Let $\mathcal{D}=\{d_1,\ldots,d_n\}$ denote a corpus of empirical studies, where each $d_i$ is a single source document with identifier $i\in\{1,\ldots,|\mathcal{D}|\}$.
We evaluate extraction and derivation over a study schema $\mathcal{S}$ (Table~\ref{tab:schema}), whose elements are \emph{properties} such as population ($P$), geolocation ($G$), sample size ($N$), statistical  method ($A$), variables ($IV$, $DV$, $V$, $S$, $U$), conditions ($C$), effect size ($E$), and measurement descriptors.

For a given document $d_i$, an \emph{atom} is a document-attributed instantiation of a schema property, written as a pair $(i,s=v)$ where $s\in\mathcal{S}$ and $v$ is the extracted value for that property in $d_i$ (e.g., $(i,N=5417)$ or $(i,A=\text{Pearson})$).
We denote the set of all atoms extractable from $d_i$ by
\[
\mathcal{A}(d_i)=\{\, (i,s=v)\ :\ s\in\mathcal{S}\,\}.
\]

Structural extraction targets not just individual atoms but \emph{bindings} of multiple atoms that must co-refer to the same underlying study statement.
We represent each bound output as a \emph{tuple} $t\in\mathcal{T}$, i.e., an ordered record
\[
t=(i; X_1,\ldots,X_k),
\]
where each $X_\ell$ is a schema-specific slot-value assignment (e.g., $P=\cdot$, $IV=\cdot$, $DV=\cdot$, $A=\cdot$, $E=\cdot$) drawn from atoms in $\mathcal{A}(d_i)$.
The arity $k$ captures how many schema fields must be jointly bound for that output.
The set of all target tuples for document $d_i$ is denoted
\[
\mathcal{I}(d_i)=\{\, t_{i1}, t_{i2}, \ldots \,\},
\]
where $t_{ij}$ indexes the $j$-th extracted tuple associated with document $d_i$.

We organize evaluation queries along two orthogonal dimensions: semantic focus and structural complexity.

\begin{table}[!htb]
\centering
\caption{Object-centric (O) list-style extraction queries. All outputs preserve the source document identifier $i$.}
\small
\begin{tabular}{p{2cm} p{3.3cm} p{9.3cm}}
\toprule
\textbf{ID} & \textbf{Tuple Pattern} & \textbf{Natural-Language Query} \\
\midrule
O\_L1\_Q1 & $(i, G)$ &
Extract the study geolocation as a country name. \\

O\_L1\_Q2 & $(i, N)$ &
Extract the reported sample size (use \texttt{null} if unavailable). \\

O\_L1\_Q3 & $(i, P)$ &
Extract the study population or unit of analysis. \\

\midrule
O\_L2\_Q1 & $(i, P, N)$ &
Extract each study population together with its sample size. \\

O\_L2\_Q2 & $(i, P, G)$ &
Extract each study population together with the study country. \\

O\_L2\_Q3 & $(i, P, G, N)$ &
Extract each study population with both country and sample size. \\
\bottomrule
\end{tabular}
\label{tab:O-list-queries}
\end{table}

\begin{table}[!htb]
\centering
\caption{Method-centric (M) list-style extraction queries. All outputs preserve the source document identifier $i$.}
\small
\begin{tabular}{p{2cm} p{3.3cm} p{9.3cm}}
\toprule
\textbf{ID} & \textbf{Tuple Pattern} & \textbf{Natural-Language Query} \\
\midrule
M\_L1\_Q1 & $(i, A)$ &
Extract the statistical method used to quantify associations. \\

M\_L1\_Q2 & $(i, V)$ &
Extract all variables. \\

M\_L1\_Q3 & $(i, IV)$ &
Extract all independent variables. \\

M\_L1\_Q4 & $(i, DV)$ &
Extract all dependent variables. \\

\midrule
M\_L2\_Q1 & $(i, V, S, U)$ &
Extract each variable together with its Scale and Unit. \\

M\_L2\_Q2 & $(i, IV, S, U)$ &
Extract each independent variable with its Scale and Unit. \\

M\_L2\_Q3 & $(i, DV, S, U)$ &
Extract each dependent variable together with its Scale and Unit. \\

M\_L2\_Q4 & $(i, IV, DV)$ &
Extract independent--dependent variable pairs. \\

M\_L2\_Q5 & $(i, IV, DV, A)$ &
Extract variable pairs together with the statistical method. \\

M\_L2\_Q6 & $(i, IV, DV, A, C, E)$ &
Extract variable pairs, statistical method, effect size (conditions). \\
\bottomrule
\end{tabular}
\label{tab:M-list-queries}
\end{table}

\noindent\textbf{Task families.}
Along the semantic dimension, we distinguish two families.
\emph{Object-centric tasks (O)} target properties of the study setting and experimental subjects (e.g., $P$, $G$, $N$), investigating identification, normalization, and correct binding under document-level attribution.
\emph{Method-centric tasks (M)} target methodological and statistical evidence structures (e.g., $A$, $V$, $IV$, $DV$, $S$, $U$, $C$, $E$), including role assignment and association tuple extraction.
To disentangle variable identification from role assignment, this family includes a role-agnostic variable inventory task that extracts $V$ as the de-duplicated union of $IV$ and $DV$ within each document.

\noindent\textbf{Query types and levels.}
We define two types of queries based on whether aggregation is performed.
\emph{List-style extraction queries} directly enumerate document-attributed atoms or bound tuples according to predefined patterns, without aggregation.
To make structural difficulty explicit, these queries are further organized into two \emph{levels} of increasing relational complexity.
L1 queries target single schema properties and require only isolated atom extraction, whereas L2 queries require binding multiple schema properties into structured tuples, involving higher-arity relational constraints (Tables~\ref{tab:O-list-queries} and~\ref{tab:M-list-queries}).

\begin{table}[!htb]
\centering
\caption{Derived statistical queries over document-attributed extraction results. 
Each query operates on specific schema atoms and applies a well-defined computation.}
\small
\begin{tabular}{p{2cm} p{3.2cm} p{9.3cm}}
\toprule
\textbf{ID} & \textbf{Target Atoms} & \textbf{Computation} \\
\midrule
O\_C\_Q1 & $(N \rightarrow \mathrm{count})$ &
Count documents with $N > 100$. \\

O\_C\_Q2 & $(N \rightarrow \mathrm{mean})$ &
Compute the mean of $N$ across all documents. \\

O\_C\_Q3 & $(N \rightarrow \mathrm{median})$ &
Compute the median of $N$ across all documents. \\

\midrule
M\_C\_Q1 & $(i, A \rightarrow \mathrm{count})$ &
For each document, count reported statistical methods. \\

M\_C\_Q2 & $(i, V \rightarrow \mathrm{count})$ &
For each document, count unique variables (union of IV and DV). \\

M\_C\_Q3 & $(i, IV \rightarrow \mathrm{count})$ &
For each document, count independent variables. \\

M\_C\_Q4 & $(i, DV \rightarrow \mathrm{count})$ &
For each document, count dependent variables. \\

M\_C\_Q5 & $(i, IV, DV, E)$ &
List $(IV, DV)$ pairs with $E > 0.7$. \\

\bottomrule
\end{tabular}
\label{tab:aggregation-queries}
\end{table}

In contrast, \emph{derived statistical queries (C)} operate on the outputs of list-style extraction.
They apply aggregation, counting, or conditional filtering to extracted atoms or tuples, yielding either corpus-level summaries or document-level derived attributes while preserving explicit document identifiers (Table~\ref{tab:aggregation-queries}).

Taken together, this formulation induces a controlled progression from single-atom extraction, to multi-atom binding, to structured statistical derivation.
Our methodological objective is to examine how model behavior changes as relational binding complexity increases and as tasks move from direct extraction to derived statistical reasoning.

\subsection{Evaluation Principles}

We evaluate model outputs using tuple-level precision, recall, and F1, following established practices in structured information extraction.

\noindent\textbf{Tuple-Level Metrics.} Given gold tuples $\mathcal{G}$ and predicted tuples $\mathcal{P}$, we compute:
$\text{Precision} = \frac{|\text{correct}|}{|\mathcal{P}|}$, $\text{Recall} = \frac{|\text{correct}|}{|\mathcal{G}|}$, and $\text{F1} = 2 \cdot \frac{\text{Precision} \cdot \text{Recall}}{\text{Precision} + \text{Recall}}$.
A predicted tuple is counted as correct only if (i) all required fields semantically match a gold tuple, and (ii) it is correctly attributed to the originating paper.

\noindent\textbf{Matching Procedure.} For each predicted tuple, we identify the best-matching gold tuple via a two-stage process:
(1) \textit{Paper constraint}: A predicted tuple may only match a gold tuple from the same source paper, enforced via citation markers (e.g., \texttt{[N]}).
(2) \textit{Semantic verification}: Candidate matches are first scored using character-level sequence similarity. Borderline candidates (similarity $< 0.95$) are then verified by an LLM judge, which determines whether the prediction and gold entry refer to the same entity. Only LLM-confirmed matches are counted as correct.
For composite tuples (e.g., \texttt{(variable, unit)}), all required fields must match independently.

\noindent\textbf{Numerical Queries.} For counting and aggregation queries (e.g., MC.1--MC.5), correctness requires exact numerical equality between the predicted value and the value computed from gold tuples.

\begin{table}[!htb]
\centering
\caption{Descriptive statistics of five domain-specific corpora, reporting summary statistics over counts of document-attributed schema atoms across domains.}
\small
\begin{threeparttable}
\begin{tabular}{p{0.9cm} p{1.1cm} | p{2.9cm} r r r r r}
\toprule
\textbf{Papers} & \textbf{Source\tnote{a}} & \textbf{Metric} & \textbf{\makecell{Sample Size}} & \textbf{\makecell{Stat. Methods}} & \textbf{\makecell{Variables}} &
\textbf{\makecell{Effect Size}} \\
\midrule
\multicolumn{7}{l}{\textbf{Civil Engineering:}  Building characteristics vs. Hotel energy use} \\
\midrule
\multirow{2}{*}{11} & \multirow{2}{*}{\cite{arenhart2024energy}} & Min/Median/Max & 6/28/51 & 1/2/5 & 2/10/32 & 2/24/46\\
 & & Total & 292 & 23 & 139 & 258 \\
\midrule
\multicolumn{7}{l}{\textbf{Medical and Health Science:} Anthropometric indicators vs. Ophthalmic outcomes} \\
\midrule
\multirow{2}{*}{9} & \multirow{2}{*}{\cite{Guan2025AnthropometricMyopia}} & Min/Median/Max & 184/5,417/1,323,052 & 1/2/3 & 2/8/16 & 16/35/78 \\
 & & Total & 1,391,776 & 18 & 71 & 328 \\
\midrule
\multicolumn{7}{l}{\textbf{Agricultural Science:} Environmental stress vs. Plant responses} \\
\midrule
\multirow{2}{*}{11} & \multirow{2}{*}{\makecell[l]{Journal\tnote{b}}} & Min/Median/Max & 8/45/232 & 1/1/6 & 4/7/11 & 4/18/54 \\
 & & Total & 871 & 20 & 72 & 226 \\
\midrule
\multicolumn{7}{l}{\textbf{Earth \& Environmental Sciences:} Carbon storage vs. Biodiversity metrics} \\
\midrule
\multirow{2}{*}{10} & \multirow{2}{*}{\cite{liu2025exploring}} & Min/Median/Max & 131/8,909/79,555 & 1/1/2 & 4/6/19 & 3/21/35 \\
 & & Total & 142,070 & 12 & 84 & 198 \\
\midrule
\multicolumn{7}{l}{\textbf{Social Science:} Compassion vs. Well-being and burnout} \\
\midrule
\multirow{2}{*}{11} & \multirow{2}{*}{\cite{Zhuniq2025well-being}} & Min/Median/Max & 37/303/1742 & 1/1/2 & 2/6/18 & 2/14/28 \\
& & Total & 4,599 & 15 & 81 & 137 \\
\bottomrule
\end{tabular}
\begin{tablenotes}
    \footnotesize
    \item[a] All papers are from meta-analysis review articles except specifically noted.
    \item[b] Retrieved \textit{Agricultural Water Management} (2005--2025) with query \textit{``(drought stress OR water stress) AND plant''}.
\end{tablenotes}
\end{threeparttable}
\label{tab:combined-corpus-stats}
\end{table}

\section{Experiments}

This section describes the experimental setup, including the corpora,
annotation protocol, preprocessing pipeline, model configurations, and input
regimes. Evaluation is performed according to the tuple-level metrics and
procedures defined in Section~\ref{sec:method}.

\subsection{Corpora and Annotation}
We construct five domain-specific corpora of primary empirical association studies spanning engineering, health, agricultural, environmental, and social sciences. The corpora are intentionally diverse in study scale, methodological complexity, and reporting practices, providing a challenging testbed for structured evidence extraction and derived statistical analysis. All included studies are primary empirical articles reporting quantitative associations with explicit statistical methods and effect-size estimates, reviews and secondary analyses are excluded.

Table~\ref{tab:combined-corpus-stats} summarizes the key descriptive statistics of the five corpora. For each domain, the table reports corpus size as well as summary statistics over counts of document-attributed schema atoms, including sample sizes, statistical methods, variables, and reported effect-size instances. These statistics highlight substantial cross-domain variation in both study scale and structural complexity of reported evidence.
Gold-standard annotations are created manually by a single expert annotator. All schema properties and association structures are extracted directly from the text, preserving explicit document identifiers for every extracted instance. Annotations include all reported variables, methods, and association estimates, regardless of statistical significance. As the task focuses on factual extraction rather than interpretive labeling, no inter-annotator agreement procedure is required. 

We release all gold-standard JSON annotations, the full query suite, experiment scripts, and the DOIs of all included articles. Due to copyright restrictions, article PDFs and their derived text representations are not redistributed, ensuring full methodological reproducibility while respecting publisher's license.

\subsection{Experimental Setup}

\noindent\textbf{PDF-to-text preprocessing.}
All source PDFs are converted to structured markdown using MinerU~2.5~\footnote{\url{https://github.com/opendatalab/MinerU}}~\cite{niu2025mineru25decoupledvisionlanguagemodel}
with default settings.
This choice is motivated by three experimental requirements: (i) faithful preservation of document structure, including multi-column layouts, tables, and mathematical expressions; (ii) retention of section hierarchy and reading order to provide models with inputs that reflect the original article organization; and (iii) deterministic, open-source preprocessing to ensure full reproducibility.
MinerU employs vision–language models for layout detection and produces markdown outputs in which tables are rendered as HTML and formulas as \LaTeX, avoiding the loss of semantically relevant content.
No manual editing, segmentation, or truncation is applied after conversion.
For each domain, all converted articles are concatenated into a single markdown input, resulting in inputs of approximately 20k tokens per domain.

\noindent\textbf{Models.}
We evaluate two state-of-the-art large language models representing complementary system classes: a closed-source model, GPT-5.2 (version 2025-12-11)~\footnote{\url{https://platform.openai.com/docs/models/gpt-5.2}}, and a large open-source vision–language model, Qwen3-VL (Qwen3-VL-235B-A22B-Instruct-FP8)~\footnote{\url{https://huggingface.co/Qwen/Qwen3-VL-235B-A22B-Instruct-FP8}}~\cite{bai2025qwen3vltechnicalreport}.
Both models are evaluated under identical decoding settings (temperature $=0.1$, default parameters).
No external tools, retrieval components, or code execution are allowed, ensuring that all results reflect the models’ intrinsic extraction and reasoning capabilities.
All inputs fall within the supported context limits of both models (Qwen3-VL: 262k tokens; GPT-5.2: 400k tokens).
Qwen3-VL is served via vLLM~\cite{vllm} on $4\times$H100 GPUs.

\noindent\textbf{Queries and input regimes.}
The same tiered query suite is applied uniformly across all corpora.
Each query is written once in natural language and reused unchanged in all experimental conditions.
For per-document evaluation, corpus-level queries are programmatically rewritten into single-document variants while preserving their semantics.
All models receive identical instructions and query formulations, ensuring strict control over prompt-level variables.
We consider two document presentation regimes.
\textbf{(1) Global input.} All article markdown files within a domain are concatenated into a single input, preceded by document identifiers and followed by the query.
\textbf{(2) Per-paper input.} Following prior work on decomposed and least-to-most prompting~\cite{zhou2023leasttomost,press-etal-2023-measuring,khot2023decomposed}, each query is applied independently to individual documents.
After per-paper extraction, a separate aggregation step uses the model to combine document-level outputs into a final answer.
No intermediate outputs are provided as context when processing subsequent documents.
Across all settings, a total of 2{,}976 LLM calls are executed.

\begin{table}[!htb]
\caption{Comparison of F1 scores by extraction group across domains, models, and input regimes (Per-Paper vs.\ Global). Column labels use abbreviated task group names (e.g., O1 = O\_L1, O2 = O\_L2, OC = O\_C, M1 = M\_L1, M2 = M\_L2, MC = M\_C).}
\centering
\label{tab:sum_results}
\small
\setlength{\tabcolsep}{4.5pt}
\resizebox{\textwidth}{!}{%
\begin{tabular}{l l | ccc ccc c || ccc ccc c}
\toprule
 & & \multicolumn{7}{c||}{\textbf{Per-Paper}} & \multicolumn{7}{c}{\textbf{Global}} \\
\cmidrule(lr){3-9} \cmidrule(lr){10-16}
 & & \multicolumn{3}{c}{Object} & \multicolumn{3}{c}{Method} & & \multicolumn{3}{c}{Object} & \multicolumn{3}{c}{Method} & \\
\textbf{Domain} & \textbf{Model} & O1 & O2 & OC & M1 & M2 & MC & All & O1 & O2 & OC & M1 & M2 & MC & All \\
\midrule
\multirow{2}{*}{Agricultural Sci.} & Qwen3-VL & .56 & .30 & .00 & .39 & .20 & .23 & .27 & .48 & .36 & .33 & .22 & .12 & .00 & .22 \\
 & GPT-5.2 & .49 & .09 & .00 & .38 & .16 & .27 & .23 & .66 & .06 & .33 & .40 & .23 & .09 & .27 \\
\midrule
\multirow{2}{*}{Social Sci.} & Qwen3-VL & .47 & .26 & .00 & .28 & .22 & .27 & .25 & .46 & .09 & .00 & .00 & .05 & .00 & .08 \\
 & GPT-5.2 & .46 & .23 & .00 & .26 & .16 & .10 & .19 & .48 & .26 & .00 & .26 & .18 & .00 & .18 \\
\midrule
\multirow{2}{*}{Medical \& Health Sci.} & Qwen3-VL & .80 & .37 & .67 & .55 & .17 & .28 & .42 & .67 & .37 & .33 & .27 & .07 & .00 & .23 \\
 & GPT-5.2 & .56 & .19 & .67 & .52 & .23 & .19 & .36 & .67 & .11 & .00 & .42 & .20 & .00 & .22 \\
\midrule
\multirow{2}{*}{Earth \& Environ. Sci.} & Qwen3-VL & .62 & .03 & .00 & .56 & .23 & .37 & .31 & .53 & .00 & .00 & .35 & .00 & .00 & .13 \\
 & GPT-5.2 & .42 & .00 & .00 & .52 & .23 & .19 & .24 & .56 & .00 & .00 & .53 & .21 & .11 & .24 \\
\midrule
\multirow{2}{*}{Civil Engineering} & Qwen3-VL & .66 & .52 & 1.0 & .48 & .27 & .26 & .47 & .76 & .55 & .33 & .21 & .01 & .01 & .24 \\
 & GPT-5.2 & .57 & .48 & .67 & .56 & .33 & .21 & .43 & .91 & .45 & .00 & .36 & .18 & .13 & .30 \\
\midrule
\multirow{2}{*}{\textit{Average}} & \textit{Qwen3-VL} & \textit{.62} & \textit{.30} & \textit{.33} & \textit{.45} & \textit{.22} & \textit{.28} & \textit{.35} & \textit{.57} & \textit{.27} & \textit{.20} & \textit{.21} & \textit{.05} & \textit{.00} & \textit{.18} \\
 & \textit{GPT-5.2} & \textit{.50} & \textit{.20} & \textit{.27} & \textit{.45} & \textit{.22} & \textit{.19} & \textit{.28} & \textit{.66} & \textit{.18} & \textit{.07} & \textit{.39} & \textit{.20} & \textit{.07} & \textit{.24} \\
\bottomrule
\end{tabular}%
}
\end{table}

\section{Results}

\subsection{Structural Trends and Domain-Level Effects}

Table~\ref{tab:sum_results} shows systematic differences across input regimes and task groups, with additional variation across domains and models.
On macro-averaged performance over all tasks, the \emph{Per-Paper} regime yields higher F1 scores than the \emph{Global} regime for both models: Qwen3-VL decreases from 0.35 to 0.18, and GPT-5.2 decreases from 0.28 to 0.24.
This pattern is observed consistently across all domains.

Across task groups, F1 scores are generally higher for single-atom extraction (L1) than for multi-atom binding (L2) under the Per-Paper regime, and this relationship holds for both object-centric and method-centric tasks.
For object-centric extraction, Qwen3-VL decreases from O1 = 0.62 to O2 = 0.30, and GPT-5.2 decreases from O1 = 0.50 to O2 = 0.20.
For method-centric extraction, both models achieve M1 = 0.45 and decrease to M2 = 0.22.

Comparing input regimes within task families, the Global regime affects L2 tasks more strongly than L1 tasks, with the largest drops observed in method-centric extraction.
For Qwen3-VL, M2 decreases from 0.22 (Per-Paper) to 0.05 (Global), and MC decreases from 0.28 to 0.00, whereas O1 decreases from 0.62 to 0.57 and O2 from 0.30 to 0.27.
For GPT-5.2, Global input yields M1 = 0.39, M2 = 0.20, and MC = 0.07, alongside O1 = 0.66 and O2 = 0.18.
In contrast, differences between Per-Paper and Global input for L1 tasks are comparatively smaller across domains.

Precision and recall both decrease as structural complexity increases, but recall more frequently reaches near-zero values first, particularly for method-centric L2 and derived tasks.
This recall-dominant degradation pattern appears consistently across models and domains.

When comparing object-centric and method-centric tasks at matched tuple arity, object-centric tasks are higher on average.
At the single-atom level (L1), macro-averaged F1 is 0.59 for O1 versus 0.38 for M1.
Within object-centric tasks, performance varies systematically by entity type: geolocation and sample size yield higher F1 scores (0.72 and 0.70, respectively), while population or unit-of-analysis extraction is substantially lower (0.35).
At the multi-atom level (L2), macro-averaged F1 remains higher for object-centric tasks (O2 = 0.24) than for method-centric tasks (M2 = 0.17).

Within method-centric extraction, task difficulty varies depending on the presence of role and relational constraints, and similar patterns recur across arity levels.
At L1, role-agnostic variable extraction $(i,V)$ achieves an F1 of 0.44, compared to 0.37 for $(i,IV)$ and 0.42 for $(i,DV)$.
At L2, the role-agnostic task $(i,V,S,U)$ reaches 0.25, whereas $(i,IV,S,U)$ and $(i,DV,S,U)$ reach 0.15 and 0.16, respectively.
For variable-pair extraction, $(i,IV,DV)$ yields an F1 of 0.26, and adding statistical method binding further reduces performance to 0.21 for $(i,IV,DV,A)$.
The same ordering is observed across both L1 and L2 tasks, with role-agnostic variants consistently achieving higher F1.
The highest-arity method-centric task $(i,IV,DV,A,C,E)$ attains an F1 of 0.00 under both input regimes and both models.

\begin{figure}[t]
    \centering
    \includegraphics[width=0.98\linewidth]{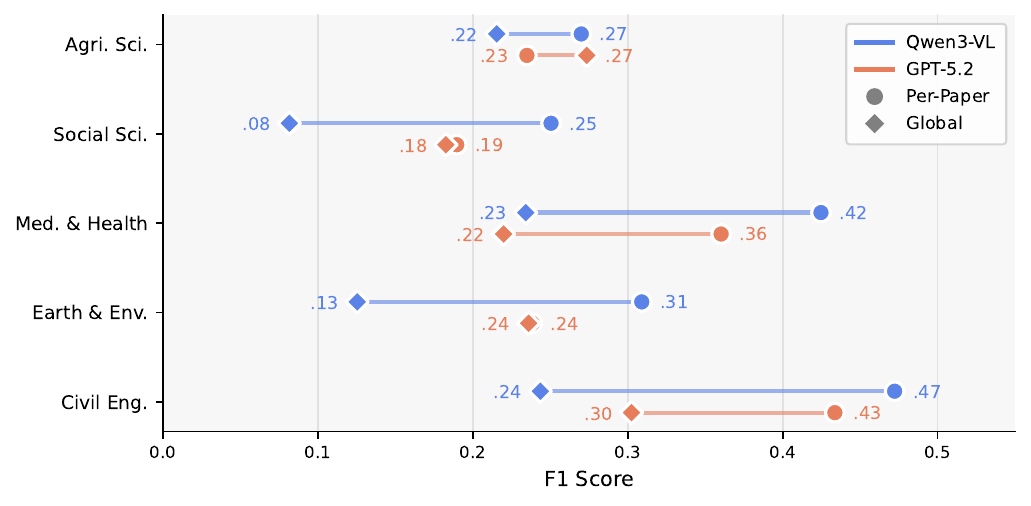}
    \caption{Domain-level comparison of macro-averaged F1 scores under the Per-Paper and Global input regimes.
    Each line connects Per-Paper and Global performance for the same domain and model, illustrating the magnitude of performance change induced by input regime variation.}
    \label{fig:pp_global_domain_drop}
\end{figure}

Domain-level differences are also visible in overall F1 scores and vary with the input regime.
Under the Per-Paper regime, overall performance across domains is relatively concentrated, whereas under the Global regime domain-level variance increases substantially.
This contrast is further illustrated by the Per-Paper → Global F1 drop across domains (Figure~\ref{fig:pp_global_domain_drop}), where all domains exhibit performance degradation but with markedly different magnitudes.
Civil Engineering achieves the highest overall performance under both regimes (Per-Paper: 0.47/0.43; Global: 0.24/0.30 for Qwen3-VL/GPT-5.2).
Medical \& Health Science ranks second under Per-Paper input (0.42/0.36), while under Global input it is close to Agricultural Science (Qwen3-VL: 0.23 vs.\ 0.22; GPT-5.2: 0.22 vs.\ 0.27).
Social Science yields the lowest overall F1, particularly under Global input (0.08/0.18).
Earth \& Environmental Science shows a large model gap under Global input, with Qwen3-VL at 0.13 and GPT-5.2 at 0.24.
These regime-dependent domain shifts motivate a closer inspection of domain-specific failure cases in the subsequent analysis.

\subsection{Error Analysis}

The quantitative patterns observed above indicate that performance degradation is not primarily attributable to failures in identifying individual entities, but instead to systematic breakdowns in preserving relational structure as task demands increase.
To better understand these trends, we analyze model outputs at the error level and identify four recurring error patterns that account for the sharp declines observed in method-centric and derived tasks.

A first prominent failure mode is \emph{role confusion in variable directionality}.
In variable-pair extraction tasks (M2.4--M2.6), 15.5\% of spurious predictions (688 out of 4,430) correspond to exact swaps of independent and dependent variables relative to the gold standard.
These errors occur despite explicit role cues in the source documents, including table headers (e.g., \emph{predictor} vs.\ \emph{outcome}), regression formulations, and methodological descriptions.
While models reliably detect the co-occurrence of relevant variables, they frequently fail to consistently bind role-specific information to those variables.
Instead, roles are often assigned based on superficial cues such as surface order of mention, indicating that role assignment remains brittle even when variable identification is correct.

Beyond local role assignment, models also exhibit systematic \emph{binding drift} across distinct analytical contexts within the same document.
Among spurious predictions in M2.5 and M2.6 under the per-paper input setting, 21.6\% (518 out of 2,403) involve variable pairs that are present in the gold annotations but are incorrectly associated with an unrelated statistical method, effect size, or experimental condition.
These errors arise most frequently in papers that reuse the same variables across multiple analyses, tables, or models.
In such cases, models successfully extract the relevant components in isolation, but fail to preserve the boundaries between separate analytical units, resulting in cross-analysis conflation.

Binding errors are further amplified in documents with high densities of reported results, leading to a \emph{multi-instance compression} effect.
For the highest-arity task (M2.6), per-paper recall decreases monotonically as the number of gold-standard tuples increases.
Papers containing 1--5 annotated instances achieve an average recall of 0.45, whereas papers with 31 or more instances drop to 0.32, with several dense papers yielding no correct extractions.
This degradation is specific to method-centric tasks, where a single document may report dozens of associations.
Object-centric queries, which typically admit only one or a small number of instances per paper, do not exhibit a comparable decline.

Finally, these upstream extraction failures propagate into pronounced \emph{error amplification} in derived statistical queries.
Aggregation tasks such as counts, means, and medians require complete and accurate upstream extractions; as a result, even limited omission or misbinding renders the final scalar output incorrect.
For example, OC.1 succeeds in only 20\% of cases despite its underlying extraction achieving an F1 of 0.66, and OC.2 fails in over 60\% of cases because excluding a single document alters the computed mean.
These results illustrate that extraction quality sufficient for qualitative inspection is inadequate for quantitative synthesis, where exhaustive recall is a strict requirement.

Taken together, these error patterns show that current models do not primarily fail at recognizing scientific entities in isolation.
Rather, failures arise from maintaining role consistency, analytical scope, and instance completeness as relational complexity increases.
These limitations directly align with the performance ceilings observed in high-arity method-centric and derived tasks, and delineate the core challenges that must be addressed for reliable meta-analytic evidence extraction.

\section{Conclusion}
We presented a structural, diagnostic study of LLM-based evidence extraction for meta-analysis, framing the problem not as an end-to-end pipeline but as a controlled progression from single-property extraction, to multi-field binding, to derived statistical computation under a fixed semantic study schema. Across five domains and two state-of-the-art models, the results draw a clear boundary between what current LLMs can do reliably and what meta-analytic workflows actually demand. Single-atom queries are often handled competently, but performance deteriorates sharply once outputs must preserve higher-arity relational structure, especially when variable roles, statistical methods, conditions, and effect sizes must be jointly bound and attributed to the correct source paper. This gap widens further in long-context, multi-document settings, where global ingestion consistently underperforms compared to per-paper decomposition, and where downstream aggregations exhibit severe error amplification, even when upstream extraction appears sufficient by conventional IE metrics.
Our error analysis shows that the dominant failures are not failures of recognition, but failures of structure: role directionality swaps, cross-analysis binding drift, instance compression in dense result sections, and cascading brittleness in numeric summaries. These patterns explain why the highest-arity association tuples collapse to near-zero performance and why seemingly simple corpus-level statistics become unreliable. The implication is that meta-analytic evidence extraction is constrained less by surface-level scientific language understanding than by the ability to maintain stable bindings over many-to-many relations and to preserve analytical scope boundaries under accumulation.
The query suite and evaluation protocol introduced here provide a principled way to measure these structural capabilities and to localize where failures begin. More importantly, they suggest a concrete agenda for the next generation of systems: schema-aware extraction that treats binding as a first-class objective, explicit representations of analytical units to prevent cross-model conflation, and verification layers that can enforce consistency between variables, roles, methods, and numeric values before aggregation. By making the structural requirements explicit and empirically mapping current failure regimes, this work offers a foundation for neural–symbolic and constraint-guided approaches that are better aligned with the precision, attribution, and completeness needs of evidence synthesis at scale.

\begin{acknowledgments}
This work was supported by the project ``HybrInt -- Hybrid Intelligence through Interpretable AI in Machine Perception and Interaction'' (Zukunft.Niedersachsen; Lower Saxony Ministry for Science and Culture; Grant ID: ZN4219). Computational resources were provided by the TIB -- Leibniz Information Centre for Science and Technology. The lead author thanks Kheir Eddine Farfar (ORKG Development Team) for support and guidance on the semantic modeling associated with the ORKG template \url{https://orkg.org/templates/R1471884} and the corresponding comparison \url{https://doi.org/10.48366/R1563383}.
\end{acknowledgments}

\section*{Declaration on Generative AI}
During the preparation of this work, the authors used ChatGPT in order to: Grammar and spelling check. After using the tool, the authors reviewed and edited the content as needed and took full responsibility for the publication’s content. 

\bibliography{ref}

\end{document}